\documentclass[opre, nonblindrev]{informs3} % current default for manuscript submission
\OneAndAHalfSpacedXI % current default line spacing

\usepackage{endnotes}
\usepackage{diagbox}
\usepackage{tikz}
\usetikzlibrary{shapes.geometric} % For geometric shapes like stars
\usetikzlibrary{calc}

\usepackage{algorithmic}

\usepackage{amsmath}
\usetikzlibrary{decorations.pathreplacing}
\usetikzlibrary{arrows.meta}
\allowdisplaybreaks
\usepackage{dsfont, pifont, mathrsfs}
\usepackage{multirow}
\usepackage{setspace}

\usepackage[hidelinks, colorlinks=true, citecolor=blue]{hyperref}
\let\footnote=\endnote

\usepackage{natbib}
 \bibpunct[, ]{(}{)}{,}{a}{}{,}%
 \def\bibfont{\small}%
 \def\bibsep{\smallskipamount}%
 \def\bibhang{24pt}%

\newcommand{\pmin}{p_{\text{min}}}
\newcommand{\pimin}{\pi_{\text{min}}}

\newcommand{\hatpsi}{\widehat{\dot{\Psi}}}

\TheoremsNumberedThrough     % Preferred (Theorem 1, Lemma 1, Theorem 2)
\ECRepeatTheorems

\EquationsNumberedThrough    % Default: (1), (2), ...
\newcommand{\var}{\operatorname{Var}}

\newcommand{\prn}[1]{\left({#1}\right)} % parentheses
\newcommand{\brk}[1]{\left[{#1}\right]} % bracket
\newcommand{\ex}[2]{\mathbb{E}_{#1}\left[#2\right]}

\newcommand{\R}{\mathbb{R}}
\newcommand{\E}{\mathbb{E}}

\renewcommand{\P}{\mathbb{P}}
\newcommand{\F}{\mathcal{F}}

\newcommand{\N}{\mathcal N}

\newcommand{\st}{\text{s.t.}}

\definecolor{arc}{RGB}{128,0,128}

\newcommand*{\circled}[1]{\lower.7ex\hbox{\tikz\draw (0pt, 0pt)%
    circle (.5em) node {\makebox[1em][c]{\small #1}};}}
\usepackage{algorithm}
\makeatletter

\begin{document}
%%%%%%%%%%%%%%%%

%\setcounter{page}{1}
%\DOI{10.1287/xxxx.0000.0000}%

\RUNAUTHOR{Ao, Chen and Simchi-Levi}


\RUNTITLE{Prediction-Guided Active Experiments}

% \TITLE{Bundling Information Goods of Decreasing Value}

\TITLE{Prediction-Guided Active Experiments}

\ARTICLEAUTHORS{%
\AUTHOR{Ruicheng Ao\textsuperscript{1} \quad\quad Hongyu Chen\textsuperscript{1} \quad\quad David Simchi-Levi\textsuperscript{1,2,3} }
\AFF{\textsuperscript{1}Institute for Data, Systems, and Society, Massachusetts Institute of Technology, Cambridge, MA 02139\texorpdfstring{\\}{}\textsuperscript{2}Department of Civil and Environmental Engineering, Massachusetts Institute of Technology, Cambridge, MA 02139\texorpdfstring{\\}{}\textsuperscript{3}Operations Research Center, Massachusetts Institute of Technology, Cambridge, MA 02139\texorpdfstring{\\}{}\EMAIL{\texttt{\{aorc, chenhy, dslevi, fengzhu\}@mit.edu}}} %, \URL{}}
} % end of the block

\ABSTRACT{%
In this work, we introduce a new framework for active experimentation, the Prediction-Guided Active Experiment (PGAE), which leverages predictions from an existing machine learning model to guide sampling and experimentation. Specifically, at each time step, an experimental unit is sampled according to a designated sampling distribution, and the actual outcome is observed based on an experimental probability. Otherwise, only a prediction for the outcome is available. We begin by analyzing the non-adaptive case, where full information on joint distribution of the predictor and the actual outcome is assumed. For this scenario, we derive an optimal experimentation strategy by minimizing the semi-parametric efficiency bound for the class of regular estimators. We then introduce an estimator that meets this efficiency bound, achieving asymptotic optimality. Next, we move to the adaptive case, where the predictor is continuously updated with newly sampled data. We show that the adaptive version of the estimator remains efficient and attains the same semi-parametric bound under certain regularity assumptions. Finally, we validate PGAE's performance through simulations and a semi-synthetic experiment using data from the US Census Bureau. The results underscore the PGAE framework’s effectiveness and superiority compared to other existing methods.
}%


%  existence of optimal policies; semi-Markov decision process; cyclic schedule}

%\KEYWORDS{}
% \HISTORY{\today.}

\maketitle
%%%%%%%%%%%%%%%%%%%%%%%%%%%%%%%%%%%%%%%%%%%%%%%%%%%%%%%%%%%%%%%%%%%%%%

% Samples of sectioning (and labeling) in OPRE
% NOTE: (1) \section and \subsection do NOT end with a period
%       (2) \subsubsection and lower need end punctuation
%       (3) capitalization is as shown (title style).
%
%\section{Introduction.}\label{intro} %%1.
%\subsection{Duality and the Classical EOQ Problem.}\label{class-EOQ} %% 1.1.
%\subsection{Outline.}\label{outline1} %% 1.2.
%\subsubsection{Cyclic Schedules for the General Deterministic SMDP.}
%  \label{cyclic-schedules} %% 1.2.1
%\section{Problem Description.}\label{problemdescription} %% 2.

% Text of your paper here

\section{Introduction}
Experimentation is fundamental to scientific research across diverse fields, from online platforms to the pharmaceutical industry, due to its capability to evaluate the performance of new treatments \citep{bakshy2014designing, cerqueira2020adaptive, li2022interference, chen2023efficient}. A central question in experimental design lies in optimizing the data collection scheme. Collecting individual data points can be costly, especially for high-quality labels, which often restricts experimenters from gathering sufficient data and can compromise statistical efficiency. At the same time, modern machine learning techniques frequently provide experimenters with a predictor of outcomes. This predictor could be a traditional machine learning model, such as Random Forests, Boosting Trees, or Neural Networks trained on historical data, or even Large Language Models (LLMs) that serve as simulators for human agents \citep{li2022interference, gui2023challenge, xie2024can}.

This study tackles the question of how existing predictors can be leveraged to create a more efficient data collection strategy. Specifically, we consider a setting in which an experimenter seeks to evaluate a new algorithm's performance with a focus on estimating the mean outcome of a known population under this new approach. This objective contrasts with traditional experimental designs, which often aim to measure treatment effects—that is, the difference in outcomes between the new algorithm (treatment) and an existing one (control) \citep{deaton2018understanding, kato24active}. Our framework assumes the experimenter possesses sufficient knowledge of control arm and is solely interested in assessing the treatment arm, which is related to clinical trials using external control data (\citealt{schmidli2020beyond}, \citealt{rahman2021leveraging}). This scenario also parallels survey sampling \citep{sarndal1980pi}, where the primary goal is to estimate the mean outcome for a specific population (e.g., income, education levels, or political views).

Our approach to experimental design considers two factors: the sampling distribution and the experimental probability. Specifically, we assume the experimenter first samples according to a designated sampling distribution, then conducts experiments and collects the outcome data based on the experimental probability; otherwise, only the predicted outcome is available. In other words, not all sampled units are tested subjected to the new algorithm. Some of the outcome of the sampled units replaced by the prediction. In this way, we can potentially minimize the number of experimental units needed and achieves a higher precision given the same budget.

Thus, the target is to optimize both the sampling distribution and experimental probability to minimize the asymptotic variance of the final estimator. This study poses two main questions: (1) How should we optimally allocate the sampling budget with known population density and an existing predictor? (2) How should we conduct inference after data collection? To address the first question, we introduce an experimental design framework called ``Prediction-Guided Active Experiments,'' which builds on a simple intuition: in a heterogeneous population, more samples should be collected from groups with either higher outcome variance or lower prediction accuracy. For the second question, we propose an estimator that achieves the lowest asymptotic variance for a given sampling scheme. This estimator is built on the recent Prediction-Powered Inference (PPI) framework (\citealt{angelopoulos2023prediction}), which has valid statistical guarantees without knowledge on the precision of the predictor.

We begin by analyzing the non-adaptive setting, where the joint distribution of individual covariates, predictions, and target outcomes is known in advance. In this scenario, we derive efficient sampling densities and experimental probabilities by minimizing the semiparametric efficiency lower bound, which is a lower bound for the asymptotic variance of any regular estimator. The resulting solution is intuitively linked to the conditional variance decomposition of the target outcome with respect to the predictor. We then consider the adaptive setting, where the relationship between predictions and actual targets is initially unknown. In this setting, we propose an adaptive scheme that aligns with the non-adaptive solution and an efficient estimator that achieves the variance lower bound in both settings. Finally, we validate our approach through simulations and semi-synthetic experiments using U.S. Census data, demonstrating the method's efficiency.

\subsection{Related Work}
Our work is related to three streams of literature: adaptive experimentation, prediction-powered inference, and active learning.

\textbf{Adaptive Experimentation.} Adaptive experimentation involves dynamically adjusting the allocation of treatment and control groups across different populations to optimize the estimation accuracy of treatment effects. The concept of adaptivity was first introduced by \cite{neyman1934two} and has been extensively developed since then (\citealt{robbins1952some}; \citealt{solomon1970optimal}; \citealt{rosenberger2001optimal}; \citealt{hahn2011adaptive}). A central principle here is that if one group (either treatment or control) demonstrates lower outcome variance, then fewer samples are required for precise estimation. With the growth of online platforms, adaptive experimentation has gained traction due to its practical applicability, as seen in recent studies (\citealt{tabord2023stratification}; \citealt{zhao2023adaptive}; \citealt{simchi2023multi}; \citealt{ao2024online}; \citealt{kato24active}). Unlike these works, our research focuses on enhancing inference by leveraging the predictive power of auxiliary models, offering a novel perspective on adaptive experimentation.

\textbf{Active Learning.} Active learning frameworks enable machine learning algorithms to selectively label specific unlabeled data points for model training, typically to maximize predictive accuracy (\citealt{settles.tr09}; \citealt{ren2021survey}). Sampling is often optimized based on uncertainty measures, with the primary goal of improving model performance (\citealt{balcan2006agnostic}; \citealt{settles2011theories}; \citealt{sener2018active}; \citealt{sinha2019variational}; \citealt{zhang2023active}). Our work aligns with active learning in that both approaches aim to optimize sampling distribution to achieve greater efficiency. However, our focus diverges in that we also incorporate the information from a predictor and the goal is to make inference on the mean outcome, while active learning predominantly aims to enhance model accuracy.

\textbf{Prediction-Powered Inference (PPI).} Our work is also inspired by The literature on prediction-powered inference (PPI) (\citealt{angelopoulos2023prediction}; \citealt{angelopoulos2023ppi++}; \citealt{zrnic2024cross}). PPI assumes access to true labels for only a subset of data, with predictions from an external model for the remainder. This framework employs established semi-parametric methods to conduct valid inference with the partially labeled data (\citealt{rubin1976inference}; \citealt{robins1994estimation}; \citealt{robins1995semiparametric}; \citealt{chernozhukov2018double}). Recent studies, such as \cite{zrnic2024active}, explore actively collecting labels to optimize statistical efficiency within the PPI framework. Our work extends this by jointly optimizing both the sampling distribution and experimental probabilities, demonstrating higher efficiency gains, as evidenced by our numerical results.

\section{Model}

\subsection{Basic Setup}

We consider a population, where each individual is characterized by a covariate $X \in \mathcal{X} \subset \mathbb{R}^d$ that represents the demographic (age, sex, occupation, etc.). The distribution of $X$ is denoted by $\mathbb{P}_X$, with a density function $q(x)$, which is assumed to be known to the experimenter. Suppose there is a new algorithm, and the goal of the experimenter is to evaluate the performance of that new algorithm on this population.  When the policy is applied to an individual with covariate $X$, the resulting outcome is denoted by $Y \in \mathbb{R}$ with the conditional distribution $\mathbb{P}_{Y \mid X}$. The conditional expectation of the outcome is given by $\mathbb{E}[Y \mid X] = \mu(X)$. The primary objective is to estimate the mean outcome, denoted by $\theta$, defined as:
\begin{equation}
\theta := \mathbb{E}[Y] = \mathbb{E}[\mathbb{E}[Y \mid X]] = \mathbb{E}_{X \sim q(x)}[\mu(X)].
\end{equation}

% Since the distribution of $X$ is known to the experimenter, the optimal way of estimating $\theta$ is by first estimating the conditional expectation $\hat\mu(X)$ for each $X$ and then aggregate them by $\hat\theta=\int_\mathcal{X}q(x)\hat\mu(x)dx$. In this way, the asymptotic variance is given by $\E[\var(Y\mid X)]$, which is smaller than $\var(Y)$, the asymptotic variance of the estimator where we take a simple average.

Now we assume that the experimenter can gather data not only by directly applying the algorithm to a sampled individual and observing the outcome $Y$, but can also by utilizing a predictor $f$ that takes $X$ and some other information $W\in\mathcal{W}\subset\R^{m}$ as input and provides predicted outcomes $F = f(X, W)\in\R$. Here, $W$ can be other characteristics of the user that is not part of the demographic, or it is some outside information like the environment when the experiments are conducted, or it can be the inherent randomness from the prediction model itself. We don't assume we know the distribution of $W$ a priori. The predictor $f$ here can be any machine learning model, such as a random forest, boosting trees, or a large language models (LLMs). 

Importantly, the  accuracy of $f$ is not assumed a priori; instead, the proposed method will adapt to the accuracy of $f$. When the predictions are accurate, the method will produce an estimator with significantly reduced error compared to traditional approaches. Even if the predictions are not highly accurate, the proposed method will still provide valid uncertainty estimates through confidence intervals.

To start with, let us first examine a naive approach. Let us say the experimenter would just sample $n$ individuals $X_1, X_2, \dots, X_n$ independently from the population distribution $\mathbb{P}_X$ and observe their outcomes $Y_1, \dots, Y_n$. Then, a simple average $\hat{\theta} = \frac{1}{n} \sum_{i=1}^n Y_i$ would yield a consistent estimator for $\theta$. However, this approach can be inefficient in two ways:

\begin{enumerate}
    \item \textbf{Sampling Distribution}: The variance of outcomes may vary significantly across subpopulations. For subpopulations with higher outcome variance, i.e., the population where the conditional mean is more difficult to analyze, it would be beneficial to sample more individuals, whereas for subpopulations with lower variance, fewer samples are needed.
    
    \item \textbf{Predictor Utilization}: The simple average does not leverage information from the predictor, which can serve as a useful surrogate to reduce the need for costly direct outcome observations. Intuitively, if we know for a specific population the estimator is very accurate, then we could reduce the sample size for that population and use the prediction to estimate the conditional mean.
\end{enumerate}

Observing the inefficiency in the above naive experimentation scheme,  we introduce a framework for \textbf{Prediction-Guided Active Experiments (PGAE)}, aimed at improving experimental efficiency in both aspects described above, for which we describle in the following section.

\subsection{Prediction-Guided Active Experiments}\label{sec:PGAE}

In this active experimentation framework, the experimenter utilizes the known covariate distribution and the prediction tool onhand to strategically select which individuals to sample and whether to apply the algorithm and observe outcomes or rely on the predictor. Specifically, at each time step $t$, the experimenter determines two key quantities: the sampling distribution $p_t(x)$ and the experiment probability $\pi_t(x)$, and the sequence of actions at each time step $t$ is as follows:

\begin{enumerate}
    \item The experimenter samples an individual $X_t$ from distribution $p_t(x)$.
    \item The experimenter observes $W_t$ and obtains the predicted outcome from the predictor, $F_t = f_t(X_t, W_t)$.
    \item With probability $\pi_t(X_t)$, the experimenter applies the algorithm and observes the actual outcome $Y_t$.
    \item The experimenter updates the sampling distribution, experimental probability, and the predictor for the next time step to $p_{t+1}$, $\pi_{t+1}$, and $f_{t+1}$, respectively.
\end{enumerate}

After $T$ time steps, the experimenter collects a dataset of the form $\{X_t, F_t, \Delta_t, \Delta_t Y_t\}_{t=1}^T$, where $\Delta_t \in \{0, 1\}$ indicates whether the experiment was conducted ($\Delta_t = 1$) or not ($\Delta_t = 0$). Then an estimation method is used at the final period to estimate the mean outcome. Importantly, compared to the naive method, the number of people that are actually subjected to the new algorithm is $\sum_{t=1}^T\Delta_t$, which is much smaller than $T$ depending on the experimental probability $\pi_t$.

The goal of this adaptive approach is to ensure that the sampling and experimentation are focused where they are most needed—sampling more from high-variance subpopulations and conducting experiments where the predictor is less reliable. This strategy helps to minimize costs while maintaining statistical validity and improving the efficiency of the estimation process. In the following section, we will derive the optimal choice of the $p_t$ and $\pi_t$ for the experimenter along with an estimation scheme that can ensure the estimator achieve the smallest asymptotic variance.

\section{The Optimal Experimental Design}
In this section, we focus on the non-adaptive setting where the experimenter chooses the sampling distribution $p(x)$ and the experimental probability $\pi(x)$ at the beginning of the experimental period. Then the experimenter sample $T$ people according to $p(x)$ and conduct actual experiments for people with covariate $x$ with probability $\pi(x)$. Under the non-adaptive setup, the final data $\{X_t, F_t, \Delta_t, \Delta_tY_t\}_{t=1}^T$ are i.i.d. for each $t$, which enables us to conduct further statistical analysis. In particular, we will first derive a variance lower bound for every regular estimator given the $T$ i.i.d. data. Then we obtain the optimal choice of $p(x)$ and $\pi(x)$ by minimizing this lower bound. 

\subsection{Efficiency Bound}
To derive the optimal choice of $p(x)$ and $\pi(x)$, we first focus on the semi-parametric efficiency bound for estimating the estimand $\theta$ given the i.i.d. data $Z_t=(X_t, F_t, \Delta_t, \Delta_tY_t)$. The efficiency bound gives the lowest asymptotic variance for a class of regular estimators (See Chapter 23.5 of \cite{van2000asymptotic} for a formal definition of regular estimators). In short, these estimators are $\sqrt{T}$-consistent estimators and remains so under any perturbation of size $1/\sqrt{T}$ of the data generating process. To start with, we first calculate the efficient influence function in below

\begin{lemma}\label{prop:efficient}
    The semi-parametric efficient influence function for $\theta$ given $\pi(x)$ and $p(x)$ and the data generation process $Z=(X, F, \Delta, \Delta Y)$ is given by 
    \begin{equation}
    \label{eq:influence}
        \dot{\Psi}(Z) = \frac{\Delta q(X)}{\pi(X)p(X)}\prn{Y-\ex{}{Y\vert X}}+\frac{(\Delta-\pi(X))q(X)}{\pi(X)p(X)}(\ex{}{Y\vert X} - \ex{}{Y\vert X,F})
    \end{equation}
\end{lemma}

The efficient influence function acts as the derivative of the final estimand with respect to the data distribution. By using the efficient influence function in Lemma \ref{prop:efficient}, we are able to calculate the variance lower bound for the regular estimators, which is simply the variance of the influence function (See Chapter 23 of \cite{van2000asymptotic}). 

\begin{theorem}\label{thm:lower_bound}
    Given the i.i.d. sampled data $\{Z_t\}_{t=1}^T$ where $Z_t=(X_t, F_t, \Delta_t, \Delta_tY_t)$ described above, for any $\sqrt{T}$-consistent regular estimator $\hat{\theta}^T$ with asymptotic variance $\sigma^2$, i.e., $\sqrt{T}(\hat\theta^T-\theta)\overset{d.}{\to} \N(0,\sigma^2)$, we have
    \begin{equation}\label{eq:lower_bound}
        \sigma^2\geq \var(\dot\Psi(Z))= \E_{X\sim p(x)}\left[ \frac{q^2(x)}{ p^2(x)}\left(\frac{1}{\pi(x)}\E[\var(Y\mid X, F)\mid X] + \var(\E[Y\mid X, F]\mid X)\right) \right] := V(p, \pi).
    \end{equation}
\end{theorem}

Theorem \ref{thm:lower_bound} provides the variance lower bound for the class of regular estimators. In particular, there are two parts in the expression, which corresponds to the  variance decomposition of $\var(Y\mid X)$ with respect to the predictor $F$. The first part involves $\E[\var(Y\mid X, F)\mid X]$, which is the the part of conditional variance $\var(Y\mid X)$ that cannot be explained by the predictor $F$. The second part relates to $\var(\E[Y\mid X, F]\mid X)$, which is the the part that can be explained by the predictor $F$. Here, the experimental probability $\pi(x)$ enters the expression only in the first term, meaning that experimentation will contribute to reducing the variance of the estimator by reducing the unpredictable part of the variance.

To present some more intuition, if $p(x)=q(x)$ and $\pi(x)=1$, which is the naive case where the experimenter samples according to the original distribution and conduct experiments on every sample, the lower bound becomes $\E[\var(Y\mid X)]$. In this case, the lower bound is attainable by first estimating the conditional mean $\E[Y\mid X]$ and then aggregating them according to the true $q(x)$. Note that for the naive estimator, the asymptotic variance is actually $\var(Y)$. The difference between $\E[\var(Y\mid X)]$ and $\var(Y)$ is because we already know the true covariate distribution, which facilitates us in estimation of the population mean. 

\subsection{Optimal Experimental Design}
Given the semi-parametric lower bound $V(p, \pi)$ for any sampling distribution $p$ and experimental probability $\pi$, we should select those $p$ and $\pi$ that minimize the lower bound. It is straightforward to see that the lower bound in \eqref{eq:lower_bound} decreases with $\pi(x)$, which is intuitive since with more actual experiments, we should be able to do estimation at a higher accuracy. Thus, we assume the experimenter faces a budget constraint $\gamma\in[0,1]$, which is the proportion of samples that are subjected to actual experimentation. In fact, for an experimental design with sampling probability $\pi(X)$, its portion of the experimental units can be calculated as $\E_{X\sim p(x)}[\pi(X)]$. Thus, given a pre-assigned experimental ratio $\gamma$, the optimization problem becomes
\begin{equation}\label{eq:opt}
\begin{aligned}
     \min_{p, \pi}&\quad V(p, \pi)\\
     \mbox{s.t.} & \quad \E_{X\sim p(x)}[\pi(X)]\leq \gamma.
\end{aligned}
\end{equation}
The solution of the optimization problem \ref{eq:opt} can be characterized in the following theorem. 
\begin{theorem}\label{thm:pi and p}
    Given the experimental ratio $\gamma$, the optimal solution $(p^*(x), \pi^*(x))$ to \eqref{eq:opt} can be characterized as
    \begin{equation}
    \begin{aligned}
        p^*(x) \propto q(x)\sqrt{\var(\E[Y\mid X, F]\mid X)}, \qquad
        \pi^*(x) \propto \gamma\sqrt{\frac{\E[\var(Y\mid X, F)\mid X]}{\var(\E[Y\mid X, F]\mid X)}}.
    \end{aligned}
    \end{equation}
    with the efficiency lower bound becomes
    \begin{equation}
        V(p^*, \pi^*) = \ex{X\sim q(x)}{\sqrt{\ex{}{\var(Y\vert X,F)\mid X}} }^2+ \ex{x\sim q(x)}{\sqrt{\var(\ex{}{Y\vert X,F}\vert X)}}^2
    \end{equation}
\end{theorem}
In Theorem \ref{thm:pi and p}, we give the relative magnitude of the absolute solution $\pi^*(x)$ and $p^*(x)$. One can obtain the actual value of $p^*$ and $\pi^*$ normalizing them according to $\int_{\mathcal{X}} p(x)dx=1$ and $\E_{X\sim p(x)}[\pi(X)]=\int_{\mathcal{X}}p(x)\pi(x)=\gamma$. If we regard $\var(\E[Y\mid X, F] \mid X)$ as the predictable variance and the $\E[\var(Y\mid X, F)\mid X]$ as the unpredictable variance. Then the sampling distribution $p^*$ is proportional to the actual density $q$ times the predictable part of the variance. As for the experimental probability $\pi$, it is proportional to the ratio between two parts --- higher unpredictable variances compared to predictable variance will lead to more experimentation in that particular group. The expression relates to the optimal sampling distribution without predictions, which is $q(x)\sqrt{\var(Y\mid X)}$ (\citealt{kato24active}). The difference is that with a predictor, the conditional variance term becomes the unpredictable component of the conditional variance. This reveals the role of experimentation in the presence of a predictor --- experimentation helps to reduce the part of variance that is unpredictable by the predictor.

\subsection{Efficient Estimator}
In this section, we derive an efficient estimator $\hat{\theta}^*$ that actually achieves the semi-parametric efficiency lower bound $V(p, \pi)$ for any given experimental design rule $p$ and $\pi$. The derivation here follows the standard one-step correction estimator in semi-parametric statistics using sample-splitting techniques (e.g. \citealt{chernozhukov2018double}, \citealt{kennedy2022semiparametric}).  Denote $Z_t = (X_t,F_t,\Delta_t,\Delta_tY_t)$ as the $t$-th sample (here we use $\Delta_tY_t$ to emphasize that we can only observe $Y_t$ while $\Delta_t=1$). Let $\hat\mu(X)$ and $\hat\tau(X,F)$ be any estimators of $\ex{}{Y\vert X}$ and $\ex{}{Y\vert X,F}$, we define the estimated influence function given $\hat\mu, \hat\tau$ for $Z_t$ as 
\begin{align*}
    \widehat{\dot{\Psi}}(Z_t;\hat\mu,\hat\tau) := \frac{q(X_t)}{\pi(X_t)p(X_t)}\brk{\Delta_t(Y_t-\hat\tau(X_t,F_t))-\pi(X_t)(\hat\mu(X_t)-\hat\tau(X_t,F_t))}.
\end{align*}
To reduce the dependence between $\hat\mu,\hat\tau$ and $\{Z_t\}_{t=1}^T$, we apply techniques of sample-splitting and cross-fitting to get efficient estimators. For samples $(Z_1,\dots,Z_T)$, we split them into $K$ disjoint folds randomly. Let $F_t\in[K]$ denote the indicator of the fold for $Z_t$ with probability $\P (F_t=k)=1/K.$ Now we define $\hat \mu^{-k}$ and $\hat\tau^{-k}$ as the estimator of $\E[Y\mid X]$ and $\E[Y\mid X, F]$ trained on observations excluding fold $k$. Our new estimator for $\theta$ is constructed as follows:
\begin{equation}
    \label{eq:crossfit1}
    \hat\theta_T := \sum_{k=1}^K\frac{N_k}{T}\brk{\ex{X\sim q(x)}{\hat\mu^{-k}(X)}+\frac{1}{N_k}\sum_{\{t:F_t=k\}}\widehat{\dot{\Psi}}(Z_t;\hat\mu^{-k},\hat\tau^{-k})}.
\end{equation}
The estimator \eqref{eq:crossfit1} consists of two parts. The first part is the standard estimator of the mean $\ex{X\sim q(x)}{\hat\mu^{-k}(X)}$, which is the population mean after obtaining the estimation of the conditional mean $\hat\mu^{-k}(X)$. Note that here, since the covariate distribution $q(x)$ is known, one can calculate the expectation here at high precision through Monte Carlo simulation. The second part is the term involving the influence function. This is the first-order correction term that makes the estimator efficient. Below, we prove that this estimator $\hat\theta_T$ is efficient under some regularity assumptions. 
% \begin{assumption}\label{assump:nonadaptive}
% \begin{enumerate}
%     \item There exist constants $\pmin,\pimin$ such that $p(x)\ge\pmin,\pi(x)\ge\pimin$ a.e..
%     \item There exist a constant $C$ such that $\max\{|Y|,|\mu|,|\hat\mu|,|\tau|,|\hat\tau|\}\le C$ a.e..
% \end{enumerate}
% \end{assumption}

\begin{theorem}
    \label{thm:efficient2}
If there exist constants $\pmin,\pimin$ and $C$ such that $p(x)\ge\pmin,\pi(x)\ge\pimin$, and  $\max\{|Y|,|\mu|,|\hat\mu|,|\tau|,|\hat\tau|\}\le C$ a.e.. Then the estimator \eqref{eq:crossfit1} is efficient, i.e.,
    \begin{align*}
        \sqrt{T}(\hat\theta_T-\theta)\overset{d}{\to}\mathcal N(0,V(p,\pi)).
    \end{align*}
\end{theorem}
Theorem \ref{thm:efficient2} holds for every sampling distribution $p$ and experimental probability $\pi$. Specifically, it holds for $p^*$ and $\pi^*$ proposed in Theorem \ref{thm:pi and p}. Combining Theorem \ref{thm:lower_bound}, \ref{thm:pi and p}, and \ref{thm:efficient2}, we know that as long as the experimenter follows the optimal $p^*$ and $\pi^*$, and uses the estimator \eqref{eq:crossfit1}, he will obtain the lowest possible asymptotic variance for estimating the population mean, which solves the experimental design problem in the non-adaptive setting.

\section{Adaptive Setting}
In this section, we consider the adaptive setting where the experimenter samples one single or a batch of data points at a time and updates the sampling and experimental probability $p_t, \pi_t$ in a sequential manner, which is the Prediction-Guided Active Experiments (PGAE) framework introduced in Section \ref{sec:PGAE}. We discuss both its implementation detail and asymptotic efficiency.

\subsection{Implementation of PGAE}\label{sec:implementation}
We first discuss how to implement PGAE in the adaptive setting, where $p_t, \pi_t$ and $f_t$ are adaptively estimated. The key is to use sampled data in two ways: (1) to estimate $\var(\E[Y\mid X, F]\mid X)$ and $\E[\var(Y\mid X, F)\mid X]$ consistently so that the sampling rule $p_t$ and $\pi_t$ converge to $p^*$ and $\pi^*$. (2) to update $f_{t}$ so that the prediction achieves a higher precision. To estimate \(\alpha(x) := \text{Var}(\mathbb{E}[Y \mid X, F] \mid X)\) and \(\beta(x) := \mathbb{E}[\text{Var}(Y \mid X, F) \mid X]\), we do the following steps:

\begin{enumerate}
    \item[\textbf{Step 1:}] \textbf{Estimate \(\mathbb{E}[Y \mid X, F]\) and \(\mathbb{E}[Y^2 \mid X, F]\):} Use data $\{X_i, F_i, \Delta_i, \Delta_iY_i\}_{i=1}^{t-1}$, selecting only points with \(\Delta_i = 1\), i.e. with actual observation. Regress \(Y\) and \(Y^2\) on \((X, F)\) with any machine learning model (e.g., linear regression, random forest) to produce estimates \(\hat{\tau}_t^{(1)}(X, F)\) and \(\hat{\tau}_t^{(2)}(X, F)\).

    \item[\textbf{Step 2:}] \textbf{Estimate \(\mathbb{E}[Y \mid X]\) and \(\mathbb{E}[(\mathbb{E}[Y \mid X, F])^2 \mid X]\):} Using \(\{X_i, \hat{\tau}_t^{(1)}(X_i, F_i)\}_{i=1}^{t-1}\), regress \(\hat{\tau}^{(1)}_t(X_i, F_i)\) and \((\hat{\tau}_t^{(1)}(X_i, F_i))^2\) on \(X_i\), denote the estimators by \(\hat{\mu}^{(1)}_t(X)\) and \(\hat{\mu}^{(2)}_t(X)\), respectively.

    \item[\textbf{Step 3:}] \textbf{Estimate \(\alpha(x)\) and \(\beta(x)\):} We can estimate \(\alpha(x)\) as \(\hat{\alpha}_t(x) = \hat{\mu}^{(2)}_t(x) - (\hat{\mu}^{(1)}_t(x))^2\).
    % $\footnote{In this step, we also truncate the estimated variance $\hat\alpha(x)$ to $[\sigma_\min, \sigma_\max]$ to ensure numerical stability}. 
    Similarly,  \(\beta(x)\) can be estimated by regressing \(\hat{\tau}^{(2)}_t(X_i, F_i) - (\hat{\tau}^{(1)}_t(X_i, F_i))^2\) on \(X_i\), for which we denote it as $\hat\beta_t(x)$.
\end{enumerate}

After obtaining the estimator $\hat{\alpha}_t(x)$ and $\hat\beta_t(x)$, one can obtain $\hat{p}_t$ and $\hat\pi_t$ as
\[\hat p_t(x) = \frac{q(x)\sqrt{\hat{\alpha}_t(x)}}{\E_{X\sim q}[\sqrt{\hat{\alpha}_t(X)}]}, \quad \hat\pi_t(x) = \min\left\{\frac{\gamma\sqrt{\hat{\beta}_t(x)}}{\ex{X\sim \hat p_t}{\sqrt{\hat{\beta}_t(X)} }  },1\right \}
% \quad \hat\pi_t(x) = \min\left\{\gamma \sqrt{\frac{\hat\alpha_t(x)}{\hat\beta_t(x)}}\bigg/\E_{X\sim q(x)}\left[ \sqrt{\frac{\hat\alpha_t(x)}{\hat\beta_t(x)}} \right], 1 \right\}\]
\]
Here, we explicitly write out the normalization term for both of the estimators. Note that in Step 3,  we also truncate the estimated variance $\hat\alpha(x)$ to $[\sigma_{\min}, \sigma_{\max} ]$ to ensure numerical stability. Thus, with the estimated $\hat{p}_t$ and $\hat\pi_t$, one can proceed the sampling phase. One can do the update in a batch manner, where the estimation is conducted after a batch of experimental units is sampled. Next, we discuss how to perform estimation of the population mean $\theta$ in the adaptive setting.

\subsection{Asymptotic Efficiency}
In the adaptive setting, all the adaptive estimators $\hat{\mu}^{(1)}_t, \hat{\mu}^{(2)}_t, \hat\tau^{(1)}_t, \hat\tau^{(2)}_t$ and $\hat\alpha_t, \hat\beta_t$ are trained use only the data in the first $t-1$ time steps, which are $\F_{t-1}$-measurable and are independent of the $t$-th observation. This property facilitate us to construct an efficient estimator as in \cite{van2008construction} and \cite{luedtke2016statistical}.
%The variance estimators $\hat\sigma_t^2(Y|X,F),\hat\sigma_t^2(\ex{}{Y|X,F}|X)$ for $\ex{}{\var(Y|X,F)|X}$ and $\var(\ex{}{Y|X,F}|X)$ are $\F_{t-1}$-measurable and truncated to interval $[\smin,\smax]$ (which means that we use $\max\{\smin,\min\{\smax,\hat\sigma_t^2(\cdot)\}\}$ to replace estimator $\hat\sigma_t^2(\cdot)$).
% The estimated scores are given by
% \begin{equation}
%     \label{eq:scores}
%     \begin{aligned}
%         \hat p_t(x)& := \frac{q(x)\hat\sigma_t(\ex{}{Y|X,F}|X)}{\ex{X\sim q(x}{\hat\sigma_t(\ex{}{Y|X,F}|X)}},\\ 
%     \hat \pi_t(x) &:= \min\{\frac{\hat\sigma_t(Y|X,F)}{\hat\sigma_t(\ex{}{Y|X,F}|X)}\frac{\ex{X\sim q(x}{\hat\sigma_t(\ex{}{Y|X,F}|X)}}{\ex{X\sim q(x)}{\hat\sigma_t(Y|X,F)}},1\}.
%     \end{aligned}
% \end{equation}
We now define the adaptive one-step correction estimator as follows.
\begin{equation}
    \label{eq:estimator_adap}
    \hat\theta_T^{Adap} := \frac{1}{T}\sum_{t=1}^T\brk{\ex{X\sim q(x)}{\hat\mu_t^{(1)}(X)}+\widehat{\dot{\Psi}}(Z_t;\hat\mu_t^{(1)}, \hat\tau_t^{(1)})},
\end{equation}
where $\widehat{\dot{\Psi}}(Z_t;\hat\mu_t^{(1)}, \hat\tau_t^{(1)})$ is given by 
\begin{align*}
    \widehat{\dot{\Psi}}(Z_t;\hat\mu_t^{(1)}, \hat\tau_t^{(1)}):=   \frac{\Delta_t q(X_t)}{\hat\pi_t(X_t)\hat p_t(X_t)}\prn{Y_t-\hat\mu_t^{(1)}(X_t)}+\frac{(\Delta_t-\hat\pi_t(X_t))q(X_t)}{\hat \pi_t(X_t)\hat p_t(X_t)}(\hat\mu_t^{(1)}(X_t) - \hat\tau_t^{(1)}(X_t,F_t)).
\end{align*}
To derive the asymptotic result for estimator $\hat{\theta}_T^{Adap}$, we will assume the adaptive estimators and the predictor converge in an appropriate sense. Moreover, we will also assume the actual distribution of $Y$ and $X$ are sub-Gaussian. Under these two regularity assumptions, we state the efficiency result for $\hat{\theta}_T^{Adap}$ as below.
% \begin{assumption}\label{assump:adap}
%     (a) Estimators $\hat{\mu}^{(1)}_t, \hat{\mu}^{(2)}_t, \hat\tau^{(1)}_t, \hat\tau^{(2)}_t$ and $f_t$ converge to the true estimand almost surely as $t\to\infty.$ (b) $Y_t-\mu(X)$ and $X_t-\ex{}{X_t}$ are zero-mean sub-Gaussian.
% \end{assumption}

\begin{theorem}
    \label{thm:efficient_adap}
    If (a) $f_t$ converge to some $f^*$ almost surely, and the estimators $\hat{\mu}^{(1)}_t, \hat{\mu}^{(2)}_t, \hat\tau^{(1)}_t, \hat\tau^{(2)}_t$ converge to their true estimand almost surely as $t\to\infty.$ (b) $Y_t-\E[Y\mid X_t]$ and $X_t-\ex{}{X_t}$ are zero-mean sub-Gaussian random variables, then we have 
    \begin{align}
        \sqrt{T}(\hat\theta_T^{Adap}-\theta)\overset{d}{\to}\N(0, V(p^*, \pi^*)).
    \end{align}
    Therefore, $\hat\theta_T$ is efficient.
\end{theorem}

To conclude this section, we summarize the complete algorithm in Algorithm \ref{alg:PGAE}.
\begin{algorithm}[h]
   \caption{Prediction-Guided Active Experimentation (PGAE)}
   \label{alg:PGAE}
\begin{algorithmic}
   \STATE {\bfseries Input:} Covariate distribution $q(x)$, initial predictor $f_1(x)$
   \STATE {\bfseries Initialization: } $\hat{p}_1 = q$, $\hat{\pi}_t = 1$
   \FOR{$t=1$ {\bfseries to} $T$}
   \STATE Sample $X_t \sim \hat{p}_t(x)$, observe $W_t$ and prediction $F_t = f_t(X_t, W_t)$
   \STATE With probability $\pi_t(X_t)$, conduct experiment and obtain $Y_t$
   \STATE Update $\hat{p}_{t+1}$ and $\hat{\pi}_{t+1}$ according to Section \ref{sec:implementation}.
   \STATE Update the predictor $f_{t+1}$ using sampled data.
   \ENDFOR
   \STATE Construct estimator $\hat{\theta}^{\text{Adap}}_T$ as in \eqref{eq:estimator_adap}
   \STATE {\bfseries Output:} $\hat{\theta}_T^{\text{Adap}}$
\end{algorithmic}
\end{algorithm}

\section{Numerical Results}
In this section, we conduct both simulation and semi-synthetic experiment to test the performance of the proposed design. In particular, we focus on the sample efficiency, meaning that given the number of units that actually received treatment, the mean squared error of different experimental scheme.
\subsection{Simulation}
\begin{figure}[h]
    \centering
    \includegraphics[width=0.95\linewidth]{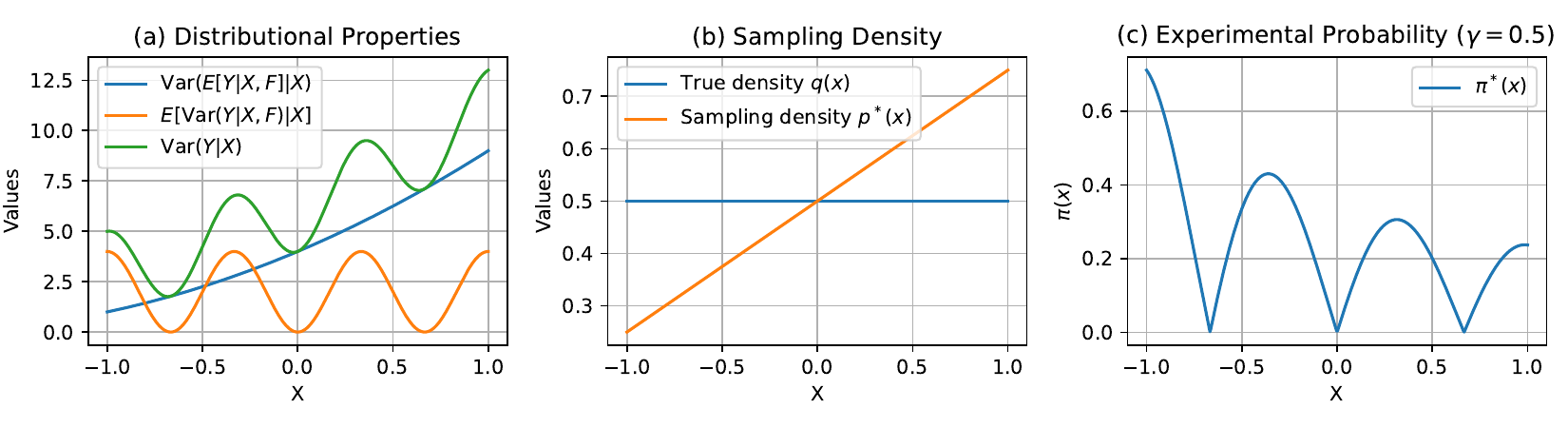}
    \caption{Simulation setup. Panel (a) is the distributional properties of $(X, F, Y)$, which includes the conditional variance $\var(Y\mid X)$ and its decomposition according to $F$. Panel (b) is the optimal sampling density $p^*$. Panel (c) is the optimal experimental probability $\pi^*$ under this setup.}
    \label{fig:simulation_setup}
\end{figure}

To start with, we first conduct a numerical simulation to validate the performance of the PGAE framework. We assume $X$ and $W$ are both real numbers sampled uniformly from $[-1,1]$, and the target $Y$ satisfies the following relationship
\[ Y =  2W +  X +  XW  + \epsilon, \]
where $\epsilon$ is a normal random noise satisfies the following condition
\[\E[\epsilon\mid W, X]= 0, \qquad \var[\epsilon\mid W, X] = 4\sin\left(\frac{3\pi X}{2}\right)^2. \]
We do not assume the predictor is perfectly accurate, instead, we only assume that the oracle estimator follows from a linear regression \[ F(X, W) =  2W +  X.\] And in the PGAE framework, we update and fit $F$ based on linear regression of $Y$ on $W$ and $X$.  Under this setup, we have $ \var(\E[Y\mid X, F]\mid X)\propto (2+X)^2$ and $ \E[\var(Y\mid X, F)] = \sin(\pi X/2)^2$, which means the optimal experimental design is given by
\[ p^*(x)\propto 2+x, \qquad \pi^*(x)\propto \left| \frac{\sin(\pi x/2)}{2+x} \right| \]
We plot this setup in Figure \ref{fig:simulation_setup}. Here, the optimal experimental design will sample more data with larger $X$, which is those data points with higher variances. The experimental probability $\pi^*$ puts more emphasis on data points with low prediction power of $F$, i.e., those points with larger $\E[\var(Y\mid X, F)\mid X]$. 

\begin{figure}[h]
    \centering
    \includegraphics[width=0.45\linewidth]{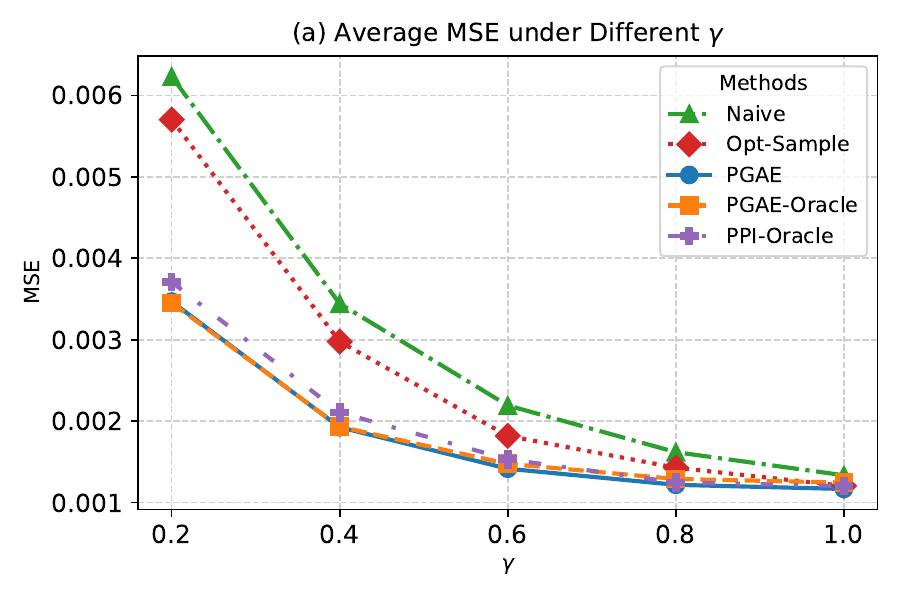}
    \includegraphics[width=0.45\linewidth]{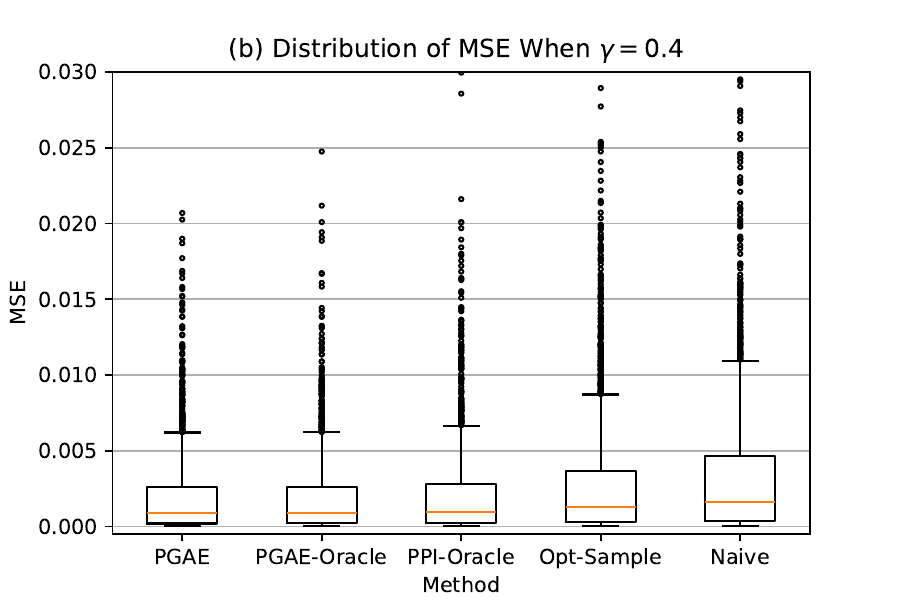}
    \caption{The averge mean square error of different estimators. Panel (a) plots the average MSE across different experimental proportion $\gamma$. Panel (b) plots the distribution of the result when $\gamma=0.4$. }
    \label{fig:simulation_result}
\end{figure}

We test in total 5 policies: (1) \textbf{PGAE}: the algorithm introduced in Algorithm \ref{alg:PGAE}. (2) \textbf{PGAE-Oracle}: Algorithm \ref{alg:PGAE} with information on the actual distribution of $(X, F, Y)$, hence knows the oracle $p^*$ and $\pi^*$. (3) \textbf{Naive}: Naive policy where data is sampled according to the true distribution $q(x)$ and estimator is constructed by sample average. (4) \textbf{Opt-Sample}: only do optimal sampling but do not use prediction in estimator. (5) \textbf{PPI-Oracle}: method proposed in \cite{zrnic2024active}, which samples according to true distribution $q$ but conduct experimentation with different probability. This method constructs estimator using the predictions according to Prediction-Powered Inference framework (\citealt{angelopoulos2023prediction}). We conduct the experiments across different experimental proportion $\gamma\in\{0.2, 0.4, 0.6, 0.8, 1.0\}$. In every experiment, the number of people that receive treatment is set to be 3000. Meaning that with predictor that uses prediction, there will be approximately $3000/\gamma$ samples. We repeat for 2000 times and plot the result in Figure \ref{fig:simulation_result}.

From the results in Figure \ref{fig:simulation_result}, we make the following observation: (1) By optimizing the sampling distribution and experimental probability, the proposed PGAE framework does achieve lower MSE across all scenarios. (2) By comparing with the Opt-Sample and PPI-Oracle estimator, we know that both the sampling density and experimental probability contributes to reduce the MSE. (3) By comparing PGAE with PGAE-Oracle, we observe that the estimation step does not reduce the efficiency of our estimator. On the contrary, it could potentially improve the accuracy. This phenomenon is seen in many semi-parametric estimators (\citealt{hitomi2008puzzling}). (4) Even if the predictor here is not perfect and the model is actually wrongly specified (we do not include $XW$ term in the regression), the estimator is still unbiased and performs relatively well. This indicates how we can harness the power of prediction in the experimentation and estimation.

\subsection{Semi-Synthetic Experiment}
We also test the algorithm on real world data from the US Census Bureau. In particular, we study the American Community Survey (ACS) Public Use Microdata Sample (PUMS), which is an annual survey that asks about education, income, citizenship, etc. We use the interface in \cite{ding2021retiring} for access of the dataset, which is also the data source in \cite{zrnic2024active}.

Here, the research question we care about is the average income of the US family in the year 2019. The target $Y$ here is the income, the covariate $X$ we choose is age and sex, other predictive variable $W$ consists of 15 features includes education, citizenship, race, etc. We conduct semi-synthetic experiments by doing a bootstrap sampling of the survey population. In particular, for each experiment, we will sample $n/10$ data points where $n=221045$ is the total number of data points. We compare the PGAE framework with three other algorithms: (1) \textbf{Naive}: naively sample according to the true distribution and output the sample average. (2) \textbf{PGAE-No-Pred}: sampling according to the optimal distribution derived in PGAE framework but only use actual experimental data for estimation. (3) \textbf{PPI}: sampling according to the true distribution, doing experiments with fixed probability $\gamma$, and construct estimator with the PPI framework (\citealt{angelopoulos2023prediction}). For a fair comparison, we fix the number of actual experimental units to be the same across different methods. That is, if the experimental portion is $\gamma$, then \textbf{Naive} and \textbf{PGAE-No-Pred} will sample $\gamma n/10$ data but \textbf{PGAE} and \textbf{PPI} will still sample $n/10$ data points with average experimental probability of $\gamma$. Here, comparison with \textbf{PGAE-No-Pred} is trying to capture the effect of using prediction in the estimator and \textbf{PPI} is trying to capture the effect of adaptive sampling.

We start with a non-adaptive approach where we use a CatBoost model trained by data from the year before (2018) as the predictor the income. Then we use that model to generate prediction for the income for each individual of the year 2019 and do not update that model. We present the MSE and width and coverage of the 95\% confidence interval in Figure \ref{fig:census_pretrain}. Here the confidence interval is constructed according to the asymptotic variance $V^*=V(p^*, \pi^*)$, where the $(1-\alpha)$-level confidence interval is given by $\hat{\theta} \pm z_{\alpha/2} \cdot \sqrt{{V^*}/{n}}$.
\begin{figure}[h]
    \centering
    \includegraphics[width=0.98\linewidth]{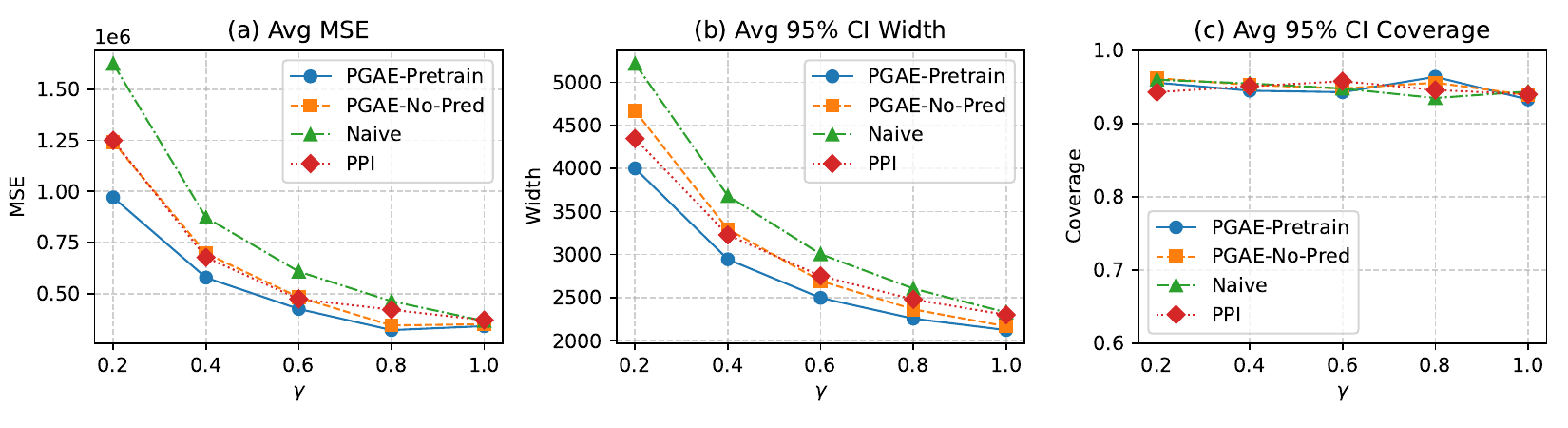}
    \caption{Results for the census data using pre-trained model. Panel (a) demonstrates the average mean square error of the predictor from 200 independent trials. Panel (b) is the width of 95\% confidence interval for three estimators. Panel (c) illustrates the coverage for three estimators.}
    \label{fig:census_pretrain}
\end{figure}

Then we test the case where we continuously update both the estimator $f_t$ and the sampling probability $p_t$ and $\pi_t$. In this case, we do not use any data from previous years and only fit the model using the current sampled data. we update the prediction model in a batch manner where we retrain the model once for every $1000$ samples added. Here, the \textbf{PPI-Adaptive} label means that the predictor is also adaptively estimated using the data of 2019. The result is in Figure \ref{fig:census_pretrain}.
\begin{figure}[h]
    \centering
    \includegraphics[width=0.98\linewidth]{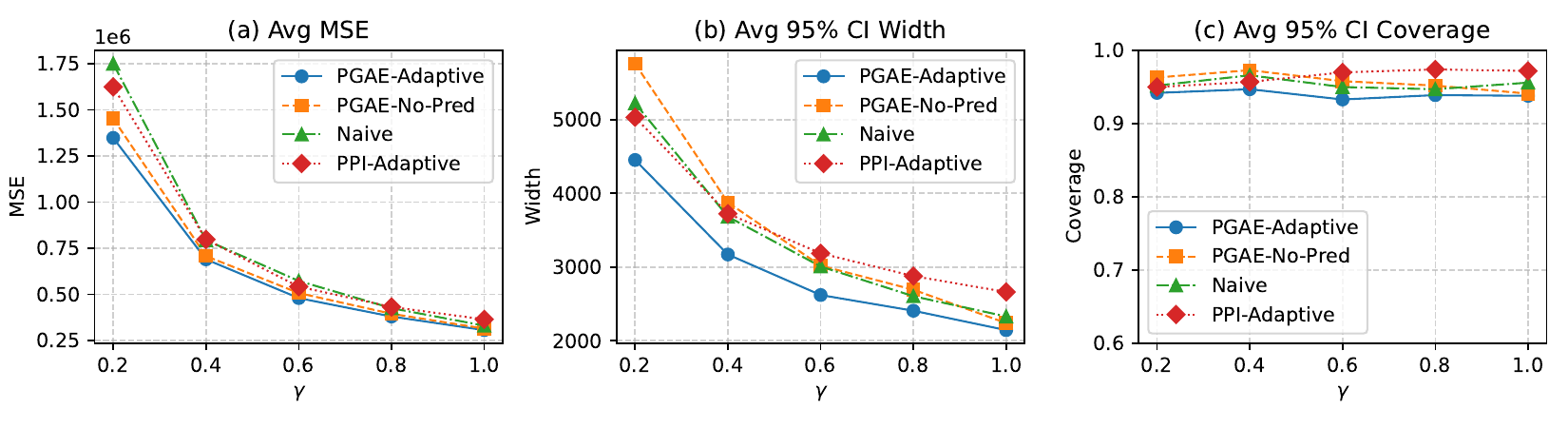}
    \caption{Results for the census data using adaptively estimated model. Panel (a) demonstrates the average mean square error of the predictor from 1000 independent trials. Panel (b) is the width of 95\% confidence interval for three estimators. Panel (c) illustrates the coverage for three estimators.}
    \label{fig:census_pretrain}
\end{figure}

From the result of both the adaptive and non-adaptive settings, we make the following observations: (1) The PGAE method outperforms all other methods in terms of the average MSE and the width of the confidence interval. The advantage is more obvious when $\gamma$ is small, i.e., when we only have limited access to actual experimental data. (2) Both the optimal sampling and optimal experimental probability contributes to the efficieny of PGAE as it outperforms both PGAE-No-Pred and PPI methods. (3) The PGAE method performs very well even if there's no pre-trained prediction model, which highlights the effectiveness when using the prediction in inference. This can be seen when we compare the performance between the two settings.

\section{Conclusion}
In this work, we propose a novel data collection scheme called \textbf{Prediction-Guided Active Experiment}, which leverages an existing predictor to guide the sampling process. This scheme prioritizes sampling data with higher outcome variance and conducts more experiments to obtain actual labels for data where the predictor underperforms. We begin by addressing the non-adaptive case, deriving a semi-parametric efficiency lower bound for any regular estimators, and identifying the optimal sampling distribution and experimental probability by minimizing this bound. We then construct an estimator whose asymptotic variance achieves the theoretical lower bound. Subsequently, we extend our approach to the adaptive case, proposing an estimator that achieves the lower bound even under adaptive sampling. Finally, we validate the performance of our proposed method through simulations and semi-synthetic experiments. We believe our method has strong potential for wide application in active experimentation, given the accessibility and ease of obtaining predictors today.

{
% \SingleSpacedXI
\bibliographystyle{informs2014}
\bibliography{main}
}

\newpage 
\setcounter{page}{1}
\begin{APPENDICES}
\small
\fontsize{9}{10}
\section{Proof of Main Results}
\subsection{Proof of Lemma \ref{prop:efficient}}
In order to derive the efficient influence function if $\Psi = \ex{X\sim q(x)}{\ex{}{Y\vert X}}$ in presence of new sampling distribution $X\sim p(x)$ and random observations $\Delta\sim \pi(x)$ given $X=x$, we can first derive the efficient influence function $\dot{\Psi}_1$ for $\Psi_1(P) = \ex{}{Y\vert X}$ for $X\sim p(x)$ (here, we emphasize that the influence function depends on the derivative related to joint distribution $\mathbb P(Y=y|X=x)$ and $p(X)$ for the constructed sampling distribution $p$), and then applying the following result in \cite{van2000asymptotic}[Ch. 25.5.3]:
\begin{equation}
    \label{eq:efficient_missing}
    \dot{\Psi}(p,\pi) = \frac{\Delta q(X)}{\pi(X)}\dot{\Psi}_1(p) - \frac{(\Delta-\pi(X))q(X)}{\pi(X)}\ex{}{\dot{\Psi}_1(p)\vert X,F}.
\end{equation}
In order to compute the efficient influence function $\dot{\Psi}_1,$ we apply a standard pathwise derivative approach (e.g. \citealt{van2000asymptotic}; \citealt{kennedy2022semiparametric}). Specifically, let $
s_\epsilon(z) = \frac{\partial}{\partial \epsilon} \log dP_\epsilon(z) \big|_{\epsilon=0} $ denote the submodel score, where $z$ stands for the vector $(x,y)$. Note that 
\begin{align*}
    \mathbb{E}\{s(Z) \mid X = x\} &= \int \frac{\partial}{\partial \epsilon} \log dP_\epsilon(z) \bigg|_{\epsilon=0} d\mathbb P(y \mid x)\\
&= \int \frac{\partial}{\partial \epsilon} \log \left\{P_\epsilon(X = x)\right\} \bigg|_{\epsilon=0} d\mathbb P(y \mid x)\\
&= \int \left\{ \frac{\partial}{\partial \epsilon} \log P_\epsilon(X = x) \bigg|_{\epsilon=0} + \frac{\partial}{\partial \epsilon} \log dP_\epsilon(y \mid x) \bigg|_{\epsilon=0} \right\} d\mathbb P(y \mid x)\\ 
&= \frac{\partial}{\partial \epsilon} \log P_\epsilon(X = x) \bigg|_{\epsilon=0}
\end{align*}
where the last equality uses the facts that \(\int dP(y \mid x) = 1\) and that scores have mean zero, i.e.,
\begin{align*}
&\int \frac{\partial}{\partial \epsilon} \log dP_\epsilon(y \mid x) \bigg|_{\epsilon=0} d\mathbb P(y \mid x) \\
&= \int \frac{\partial}{\partial \epsilon} \frac{dP_\epsilon(y \mid x)}{dP(y \mid x)} \bigg|_{\epsilon=0} d\mathbb P(y \mid x)\\
&= \int \frac{\partial}{\partial \epsilon} dP_\epsilon(y \mid x) \bigg|_{\epsilon=0} \\
&= \frac{\partial}{\partial \epsilon} \int dP_\epsilon(y \mid x) \bigg|_{\epsilon=0} \\
&= 0
\end{align*}
where the first equality used the fact that \(\frac{\partial}{\partial \epsilon} \log dP_\epsilon(y \mid x) = \frac{\partial}{\partial \epsilon} dP_\epsilon(y \mid x)/dP_\epsilon(y \mid x)\). Therefore, the pathwise derivative for $\frac{\partial}{\partial\epsilon}\Psi_1(P)\big\vert_{\epsilon=0}$ is given by
\begin{align*}
&\frac{\partial}{\partial \epsilon} \int y \, dP_\epsilon(y \mid x) \bigg|_{\epsilon=0} \\
&= \int y \left\{ \frac{\partial}{\partial \epsilon} \log dP_\epsilon(y \mid x) \right\} \bigg|_{\epsilon=0} d\mathbb P\mathbb (y \mid x) \\
&= \int y \left\{ \frac{\partial}{\partial \epsilon} \log \frac{dP_\epsilon(z)}{P_\epsilon(X = x)} \right\} \bigg|_{\epsilon=0} d\mathbb P(y \mid x) \\
&= \int y \left\{ \frac{\partial}{\partial \epsilon} \log dP_\epsilon(z) - \frac{\partial}{\partial \epsilon} \log P_\epsilon(X = x) \right\} \bigg|_{\epsilon=0} dP_\epsilon(y \mid x) \\
&= \mathbb{E}[ Y s_\epsilon(Z) \mid X = x ] - \mathbb{E}[ s_\epsilon(Z) \mid X = x ] \mathbb{E}[Y \mid X = x],
\end{align*}
where the first inequality comes from exchanging integrals and derivatives due to $P_{\epsilon=0}=\mathbb P$. On the other hand, for $\dot{\Psi}_1(p) = \frac{\mathds{1}\{X=x\}}{p(x)}(Y-\ex{}{Y\vert X=x}),$ we have 
\begin{align*}
    &\int \dot{\Psi}_1(p)s_\epsilon(z)d\mathbb P(z)\\
    &= \ex{}{\frac{\mathds 1\{X=x\}}{p(x)}(Y-\ex{}{Y\vert X=x})s_\epsilon(Z)}\\ 
    &= \mathbb{E}[ Y s_\epsilon(Z) \mid X = x ] - \mathbb{E}[ s_\epsilon(Z) \mid X = x ] \mathbb{E}[Y \mid X = x],
\end{align*}
which shows that the efficient influence function for $\Psi_1(p)$ is given by
\begin{align*}
    \dot{\Psi}_1(p) = \frac{\mathds 1\{X=x\}}{p(x)}(Y-\ex{}{Y\vert X=x}).
\end{align*}
Now, since $q(x)$ is given as prior, this is deterministic function throughout the process without need for estimating. For $\Psi = \ex{X\sim q(x)}{\ex{}{Y\vert X}}$, combining the above equation with \eqref{eq:efficient_missing} leads to 
\begin{equation*}
    \dot{\Psi}(p,\pi) = \frac{\Delta q(X)}{\pi(X)p(X)}\prn{Y-\ex{}{Y\vert X}}+\frac{(\Delta-\pi(X))q(X)}{\pi(X)p(X)}(\ex{}{Y\vert X} - \ex{}{Y\vert X,F}).
\end{equation*}

\subsection{Proof of Theorem \ref{thm:lower_bound}}
By Proposition \ref{prop:efficient}, it is sufficient to show that 
\begin{align*}
V(p,\pi) = \var\brk{\dot{\Psi}(p,\pi))}&= \var\brk{\frac{\Delta q(X)}{\pi(X)p(X)}\prn{Y-\ex{}{Y\vert X}}+\frac{(\Delta-\pi(X))q(X)}{\pi(X)p(X)}(\ex{}{Y\vert X} - \ex{}{Y\vert X,F}) }    \\ 
& =  \var\brk{\frac{\Delta q(X)}{\pi(X)p(X)}\prn{Y-\ex{}{Y\vert X,F}}-\frac{q(X)}{p(X)}(\ex{}{Y\vert X} - \ex{}{Y\vert X,F}) } 
\end{align*} .
Define $A = \frac{\Delta q(X)}{\pi(X)p(X)}\prn{Y-\ex{}{Y\vert X,F}}$ and $B = \frac{q(X)}{p(X)}(\ex{}{Y\vert X} - \ex{}{Y\vert X,F}).$ Then $var\brk{\dot{\Psi}(p,\pi))} = \var[A]+\var[B]-2\text{Cov}(A,B).$
For Var[$A \mid X$], we have

$$
\text{Var}[A \mid X] = \left(\frac{q(X)}{\pi(X) p(X)}\right)^2 \text{Var}\left[\Delta \left(Y - \mathbb{E}[Y \mid X, F]\right) \mid X\right].$$

Since $\Delta \text{ and } Y - \mathbb{E}[Y \mid X, F]$ are independent given $X$:
$$
\text{Var}\left[\Delta \left(Y - \mathbb{E}[Y \mid X, F]\right) \mid X\right] = \pi(X) \mathbb{E}\left[\text{Var}(Y \mid X, F) \mid X\right].$$

Therefore,
$$
\text{Var}[A \mid X] = \frac{q^2(X)}{\pi(X) p^2(X)} \mathbb{E}\left[\text{Var}(Y \mid X, F) \mid X\right].
$$
Then we have
$$
\text{Var}[A] = \mathbb{E}_{X\sim q(x)}\left[\text{Var}[A \mid X]\right] = \mathbb{E}_{X\sim q(x)}\left[\frac{q^2(X)}{\pi(X) p^2(X)} \mathbb{E}\left[\text{Var}(Y \mid X, F) \mid X\right]\right].
$$
% For $A,$ we have 
% \begin{align*}
%     \var[A|X] = \prn{\frac{q(X)}{\pi(X)p(X)}}^2\var\brk{\Delta(Y-\ex{}{Y|X}|X }.
% \end{align*}
% Since $\Delta$ and $Y$ are independent given $X$, we have
% \[ 
% \text{Var} \left[\Delta \left(Y - \mathbb{E}[Y \mid X]\right) \mid X \right] = \pi(X) \text{Var}(Y \mid X).
% \]
% Therefore,
% $$
% \text{Var}[A \mid X] = \left(\frac{q(X)}{\pi(X) p(X)}\right)^2 \pi(X) \text{Var}(Y \mid X) = \frac{q^2(X)}{\pi(X) p^2(X)} \text{Var}(Y \mid X).
% $$
% It follows that
% $$
% \text{Var}[A] = \mathbb{E}_{X\sim q(x)} \left[\text{Var}[A \mid X]\right] = \mathbb{E}_{X\sim q(x)} \left[\frac{q^2(X)}{\pi(X) p^2(X)} \text{Var}(Y \mid X)\right].
% $$
For Var$(B)$, we have
$$
\text{Var}[B] = \left(\frac{q(X)}{p(X)}\right)^2 \text{Var}\left(\mathbb{E}[Y \mid X] - \mathbb{E}[Y \mid X, F]\right) = \left(\frac{q(X)}{p(X)}\right)^2 \text{Var}\left(\mathbb{E}[Y \mid X, F] \mid X\right).
$$
Then we have
$$
\text{Var}[B] = \mathbb{E}_{X\sim q(x)}\left[\left(\frac{q(X)}{p(X)}\right)^2 \text{Var}\left(\mathbb{E}[Y \mid X, F] \mid X\right)\right].
$$
For the covariance term \text{Cov}[A, B]:
\[
\text{Cov}[A, B] = -\frac{q^2(X)}{\pi(X) p^2(X)} \mathbb{E}\left[\Delta \left(Y - \mathbb{E}[Y \mid X, F]\right)\left(\mathbb{E}[Y \mid X] - \mathbb{E}[Y \mid X, F]\right) \mid X\right].
\]

Since $ \Delta$ is independent of $Y$ and $F$ given $X$, for the correlation we have:
\[
\mathbb{E}\left[\Delta \left(Y - \mathbb{E}[Y \mid X, F]\right)\left(\mathbb{E}[Y \mid X] - \mathbb{E}[Y \mid X, F]\right) \mid X\right] = \pi(X) \cdot 0 = 0.
\]
Therefore
\[
\text{Cov}[A, B] = 0.
\]
Combining the results above, we get
\begin{align*}
   \var\brk{\dot{\Psi}(p,\pi))} &= \var[A]+\var[B]-2\text{Cov}(A,B)\\ 
   &=\mathbb{E}_{X\sim q(x)} \left[\frac{q^2(X)}{\pi(X) p^2(X)} \mathbb{E}\left[\text{Var}(Y \mid X, F) \mid X\right] + \left(\frac{q(X)}{p(X)}\right)^2 \text{Var}\left(\mathbb{E}[Y \mid X, F] \mid X\right)\right]=V(p,\pi).
\end{align*}

\subsection{Proof of Theorem \ref{thm:pi and p}}
By Theorem \ref{thm:lower_bound}, we aim to solve the following constraint optimization problem:
\begin{equation}\label{eq:prob_minimize}
    \begin{aligned}
        \min_{p,\pi}\qquad&\E_{X\sim p(x)}\left[ \frac{q^2(x)}{\pi(x) p^2(x)}\E[\var(Y\mid X, F)\mid X] + \frac{q^2(x)}{p^2(x)}\var(\E[Y\mid X, F]\mid X) \right],\\ 
        \st \qquad & \int p(x)dx = 1\\ 
        & \ex{X\sim p(x)}{\pi(X)} \le \gamma
    \end{aligned}
\end{equation}
Denote by $A(x):=\E[\var(Y\mid X, F)\mid X=x]$ and $\var(\E[Y\mid X, F]\mid X=x)$. The Lagrangian multiplier augmented Lagrangian functional is then given by
\begin{align*}
    L(p,\pi,\lambda,\mu) = \int\prn{\frac{q^2(x)A(x)}{\pi(x)p(x)}+\frac{q^2(x)B(x)}{p(x)}-\lambda p(x)+\mu p(x)\pi(x)}dx + \lambda + \mu\gamma.
\end{align*}
We can then compute the corresponding functinoal derivatives:
\begin{align*}
    \frac{\delta L}{\delta p(x)}& = -\frac{q^2(x)A(x)}{\pi(x)p^2(x)} - \frac{q^2(x)B(x)}{p^2(x)}-\lambda +\mu \pi(x), \\
    \frac{\delta L}{\delta \pi(x)}& = -\frac{q^2(x)A(x)}{\pi^2(x)p(x)}+\mu p(x).
\end{align*}
By stationarity conditions:
\begin{align*}
    \frac{\delta L}{\delta p(x)} = 0, \quad \frac{\delta L}{\delta\pi(x)} = 0.
\end{align*}
Solve the equation for $\pi(x)$, we get
\begin{align*}
    \mu p(x) = \frac{q^2(x)A(x)}{\pi^2(x)p(x)},\Rightarrow \mu p^2(x) = \frac{q^2(x)A(x)}{\pi^2(x)}.
\end{align*}
Substituting the above equation into the derivative with respect to $p(x),$ we have 
\begin{align*}
    \frac{-q^2(x)A(x)}{\pi(x)p^2(x)}-\frac{q^2(x)B(x)}{p^2(x)}-\lambda+\frac{q^2(x)A(x)}{\pi^2(x)p(x)} = 0,
\end{align*}
which simplifies to 
\begin{align*}
    \lambda = -\frac{q^2(x)B(x)}{p^2(x)}.
\end{align*}
Since $\lambda,\mu$ are fixed for all $x$ and all constraints need to be satisfied, we get 
\begin{align*}
    p(x)\propto q(x)\sqrt{B(x)}
\end{align*}
and 
\begin{align*}
    \pi(x) \propto \gamma\sqrt{\frac{A(x)}{B(x)}}.
\end{align*}
Since $\mu\ge 0$ from the deduction above, by \citealt{gelfand2000calculus} and \citealt{rockafellar2009variational}, the above solutions are optimal for solving \eqref{eq:prob_minimize}.

\subsection{Proof of Theorem \ref{thm:efficient2}}\label{sec:proof_efficient}
In this section, we use $P$ to represent the true joint distribution of $Z$. We use $P_T$ to denote the operator of the empirical estimator sampled from $P$ with sample size $T$. For an estimated version $\hat P$, given by some empirical distribution (e.g., $P_T$ and the corresponding estimates $\hat\mu$ and $\hat\tau$), we make a slight abuse of notation by using $\widehat{\dot{\Psi}}(Z; \hat P)$ to represent $\widehat{\dot{\Psi}}(Z; \hat\mu, \hat\tau)$. We begin with the von Mises expansion (see, e.g., \citealt{kennedy2022semiparametric}) of $\Psi = \mathbb{E}_{x\sim q(x)}[\mathbb{E}[Y\vert X]]$:
\begin{equation}
    \label{eq:expansion}
    \Psi(\hat P)-\Psi(P) = \int \dot\Psi(P)d(\hat P-P)(z) + R_2(\hat P,P),
\end{equation}
% where $P$ and $\Psi(P)$ represent the joint distribution of $X,Y,\Delta$ and the corresponding efficient influence function. $\hat P$ can represent an estimated distribution. $R_2(\hat P,P)$ is a second-order remainder term. For the estimator $\hat P$ given by empirical distribution $P_T$ of $p$ and estimated $\hat\mu,\hat\tau$, with a slight abuse of notations we use $P_T(\cdot)$ to denote the operator of empirical estimator of $\cdot$ and use $\widehat\dot{\Psi}(Z;\hat P)$ to represent $\widehat{\dot{\Psi}}(Z;\hat\mu,\tau).$ 
Then we have decomposition
\begin{equation}
    \label{eq:decomp1}
    \begin{aligned}
        \hat\theta_T - \theta &= \sum_{k=1}^K\frac{N_k}{T}\brk{\mathbb{E}_{X\sim q(x)}[\hat\mu^{-k}(X)] + \frac{1}{N_k} \sum_{t=1}^T \widehat{\dot{\Psi}}(Z_t; \hat\mu^{-k}, \hat\tau^{-k}) - \mathbb{E}_{X\sim q(x)}[\mu(X)]}\\
        & = \sum_{k=1}^K\frac{N_k}{T}\brk{\Psi(\hat P_T^k) + \hat P_T^k[\widehat{\dot{\Psi}}(Z;\hat P_T^{-k})] - \Psi(P)}\\
        & = \sum_{k=1}^K\frac{N_k}{T}\brk{\Psi(\hat P_T^k) - \Psi(P) + \hat P_T^k[\widehat{\dot{\Psi}}(Z;\hat P_T^{-k})]}\\
        & = \sum_{k=1}^K\frac{N_k}{T}\brk{\sum_{k=1}^K\frac{N_k}{T}(\hat P_T^k - P)(\widehat{\dot{\Psi}}(Z;\hat P_T^{-k})) + R_2(\hat P_T^k, P)}\\
        % & = (\hat P_T^k - P)(\dot\Psi(Z;P)) + (P_T - P)(\widehat{\dot{\Psi}}(Z;\hat P_T^{-k}) - \Psi(Z;P)) + R_2(\hat P_T^k, P)\\
        &=\sum_{k=1}^K\frac{N_k}{T}\brk{\underset{:=S^k}{\underbrace{(\hat P_T^k - P)(\dot\Psi(Z;P))}} + \underset{:=T_1^k}{\underbrace{(\hat P_T^k - P)(\widehat{\dot{\Psi}}(Z;\hat P_T^{-k}) - \Psi(Z;P))}} + \underset{:=T_2^k}{\underbrace{R_2(\hat P_T^k, P)}}}.
        % & \overset{\Delta}{=} S^k + T_1^k + T_2^k.
    \end{aligned}
\end{equation}
The first term $S^k=(\hat P_T^k-P)(\dot\Psi(Z;P))$ is a simple sample average of a fixed function. Therefore, by the central limit theorem, we have that $S^k$ approaches normal distribution $\mathcal N(0,\var(K\Psi(Z;P))/T)$ up to error of $o_{\mathbb P}(K/T).$

For the second term $T_1^k=(\hat P_T^k - P)(\widehat{\dot{\Psi}}(Z;\hat P_T^{-k}) - \dot\Psi(Z;P)),$ which is called an empirical process term. We have 
\begin{align*}
    T_1^k = &(\hat P_T^k-P)\left[\frac{q(X)}{\pi(X)p(X)}\brk{\Delta(Y-\hat\tau^{-k}(X,F))-\pi(X)(\hat\mu^{-k}(X)-\hat\tau^{-k}(X,F))}\right.\\
    &\left.\quad-\frac{q(X)}{\pi(X)p(X)}\brk{\Delta(Y-\tau(X,F))+\pi(X)(\mu(X)-\tau(X,F))}\right]\\ 
    =& (\hat P_T^k-P)\left[\frac{q(X)}{\pi(X)p(X)}[\Delta(\tau-\hat\tau^{-k})(X,F)+\pi(X)[(\mu-\hat\mu^{-k})(X)-(\tau-\hat\tau^{-k})(X,F)]\right].
\end{align*}
Under Assumption \ref{assump:nonadaptive}, we have that
\begin{align*}
    &\left\vert\frac{q(X)}{\pi(X)p(X)}[\Delta(\tau-\hat\tau^{-k})(X,F)+\pi(X)[(\mu-\hat\mu^{-k})(X)-(\tau-\hat\tau^{-k})(X,F)]\right\vert
    \\&\le  \frac{2}{\pmin\pimin}(|\tau(X,F)-\hat\tau^{-k}(X,F)|+|\mu(X)-\hat\mu^{-k}(X)|).
\end{align*}
Note that the estimators $\hat\mu^{-k},\hat\tau^{-k}$ are independent of $\hat P_T^k$ and converges to $\mu,\tau$ in distribution as $T\to\infty.$ We know that $T_1^k = o_{\P}(1/\sqrt{N_k})= o_{\P}(K/\sqrt{T}).$
% Therefore, the term $T_1^k$ is bounded by 
% \begin{equation*}
%     \label{eq:T1_bound}
%     |T_1^k| \le 2\pmin^{-1}\pimin^{-1}\norm{\hat P_T^k-P}(\norm{\tau-\hat\tau^{-k}}+\norm{\mu-\hat\mu^{-k}}).
% \end{equation*}
% It follows  that $T_1^k=O_{\mathbb P}((\norm{\tau-\hat\tau^{-k}}+\norm{\mu-\hat\mu^{-k}})/\sqrt{n})$.

Note that
\[
\dot{\Psi}(P) = \frac{\Delta q(X)}{\pi(X)p(X)}\prn{Y-\ex{}{Y\vert X}}+\frac{(\Delta-\pi(X))q(X)}{\pi(X)p(X)}(\ex{}{Y\vert X} - \ex{}{Y\vert X,F}).
\]
We can directly get that 
\begin{align*}
    T_2^k &= R_2(\hat P_T^k,P) \\
    % &= \int\brk{\frac{1}{\hat\pi(x)\hat P_T^k(x)}-\frac{1}{\pi(x)p(x)}}\brk{\mu(x)-\hat\mu(x)}\pi(x)p(x)q(x)dx\\ 
    % &= \int \frac{p(x)q(x)}{\hat\pi(x)\hat P_T^k(x)}\brk{\pi(x)(\mu(x)-\hat\mu(x))+(\pi(x)-\hat\pi(x))(\hat\mu(x)-\hat\tau^{-k}(x,F)} dx + \int q(x)(\hat\mu(x)-\mu(x))dx.
    &= \int q(x)(\mu(x)-\hat\mu^{-k}(x)) dx + \int q(x)(\hat\mu^{-k}(x)-\mu(x))dx\\ 
    & = 0.
\end{align*}

Combining the results above, we arrive at 
\begin{align*}
    \sqrt{T}(\hat\theta_T-\theta) = \sqrt{T}\sum_{k=1}^K\frac{N_k}{T}S^k + o_{\P}(1).
\end{align*}
Since $\sqrt{T}\frac{N_k}{T}S^k\overset{d}{\to} \mathcal N(0,\frac{K}{T}\var(\dot{\Psi}(P))$, we know that 
\begin{align*}
    \sqrt{T}(\hat\theta_T-\theta) \overset{d}{\to}\mathcal N(0,\var(\dot{\Psi}(P))
\end{align*}
and the estimator $\hat\theta_T$ is efficient.

% For $B$, the conditional variance is given by
% $$
% \text{Var}[B \mid X] = \left(\frac{q(X)}{\pi(X) p(X)}\right)^2 \text{Var}\left[\left(\Delta - \pi(X)\right)\left(\mathbb{E}[Y \mid X] - \mathbb{E}[Y \mid X, F]\right) \mid X\right].
% $$
% Since $\Delta - \pi(X) \text{ and } \mathbb{E}[Y \mid X] - \mathbb{E}[Y \mid X, F]$ are uncorrelated given $X$:
% $$
% \text{Var}[B \mid X] = \left(\frac{q(X)}{\pi(X) p(X)}\right)^2 \pi(X) (1 - \pi(X)) \text{Var}\left(\mathbb{E}[Y \mid X, F] \mid X\right).
% $$
% Then we can get the variance as
% $$
% \text{Var}[B] = \mathbb{E}_{X\sim q(x)} \left[\text{Var}[B \mid X]\right] = \mathbb{E}_{X\sim q(x)} \left[\left(\frac{q(X)}{\pi(X) p(X)}\right)^2 \pi(X)(1 - \pi(X)) \text{Var}\left(\mathbb{E}[Y \mid X, F] \mid X\right)\right].
% $$
% For the covariance of $A$ and $B$, since $\Delta $ is independent of $Y,F$ given $X$, we have $\text{Cov}(A,B|X)=0$. Therefore, $\text{Cov}(A,B) = \ex{X}{\text{Cov}(A,B|X)}=0.$
% For the covariance of $A$ and $B$, note that $\ex{}{B}=0$. Then we have
% \[ 
% \mathbb{E}[AB \mid X] = \left(\frac{q(X)}{\pi(X) p(X)}\right)^2 \mathbb{E}\left[\Delta \left(\Delta - \pi(X)\right)\left(Y - \mathbb{E}[Y \mid X]\right)\left(\mathbb{E}[Y \mid X]\left(\mathbb{E}[Y \mid X] - \mathbb{E}[Y \mid X, F]\right)\right) \mid X\right].
% \]
% Since 

\subsection{Proof of Theorem \ref{thm:efficient_adap}}
Define
\begin{align*}
    \xi_t = (\kappa^*)^{-1/2}\brk{\ex{X\sim q(x)}{\hat\mu_t^{(1)}(X)}+\widehat{\dot{\Psi}}(Z_t;\hat P_t)-\theta},
\end{align*}
where 
\begin{align*}
    \kappa^* = \ex{X\sim p^*(x)}{\frac{(Y-\mu(X))^2q(x)^2}{\pi^*(X) p^*(X)^2}+\frac{(1-\pi^*(X))q(X)^2}{\hat p^*(X)^2}(\mu(X)-\tau(X,F))^2}
\end{align*}
We have the following two lemmas.
\begin{lemma}
    \label{lem:martingale}
    The sequence $\{\xi_t\}_{t=1}^T$ is martingale subject to distribution $\{\hat P_t\}_{t=1}^T$, i.e.
    \begin{align*}
        \ex{\hat P_t}{\xi_t|\F_{t-1}}:=\ex{Y_t\sim \P(y|X_t),X_t\sim \hat p_t(x),\Delta_t\sim \hat\pi_t(x)}{\xi_t|\F_{t-1}}=0.
    \end{align*}
\end{lemma}
\begin{lemma}[\cite{hamilton2020time} Prop 7.9]
    \label{lem:clt}
    Suppose $\{W_t\}_{t=1}^\infty$ is a martingale sequence. Let $\bar W_T=\frac{1}{T}\sum_{t=1}^TW_t.$ Suppose that 
    \begin{itemize}
        \item [(a)] $\ex{}{W_t^2}=\sigma^2 >0$ and $\frac{1}{T}\sum_{t=1}^T\sigma_t^2\to\sigma^2>0$.
        \item [(b)]$ \ex{}{|W_t|^r} <\infty$ for some $r>2$,$\forall t\in\mathbb N$.
        \item [(c)] $\frac{1}{T}\sum_{t=1}^TW_t^2\overset{p}{\to}\sigma^2.$
    \end{itemize}
    Then it holds that $\sqrt{T\bar W_T}\overset{d}{\to}\N(0,\sigma^2).$
\end{lemma}
With Lemma \ref{lem:martingale} and \ref{lem:clt}, it suffices to show that the three conditions in Lemma \ref{lem:clt} holds for sequences $\{\xi_t\}_{t=1}^T.$ Then by Lemma \ref{lem:clt}, we know that $\sqrt{T}(\hat\theta-\theta)\overset{d}{\to}\N(0,\var(\Psi))$ and hence the estimator $\hat\theta_T$ is efficient. The proof of Lemma \ref{lem:martingale}
is left to \ref{sec:proof_lemma}.
\paragraph*{\textbf{Verification of condition (a)}} 
We first show that $\sum_{t=1}^T\ex{\hat P_t}{\xi_t^2|\F_{t-1}}\overset{P}{\to} 1.$ We first split the term $\ex{\hat P_t}{\xi_t^2|\F_{t-1}}$ into three parts.
\begin{align*}
    &\ex{\hat P_t}{\brk{\ex{X\sim q(x)}{\hat\mu_t^{(1)}(X)}+\widehat{\dot{\Psi}}(Z_t;\hat P_t)-\theta}^2|\F_{t-1}} \\ 
    % &=\ex{\hat P_t}{\widehat{\dot{\Psi}}(Z_t;\hat P_t)^2|\F_{t-1}}+ \ex{\hat P_t}{(\ex{X\sim q(x)}{\hat\mu_t^{(1)}(X)}-\theta)^2|\F_{t-1}} + 2\ex{\hat P_t}{\widehat{\dot{\Psi}}(Z_t;\hat P_t)(\ex{X\sim q(x)}{\hat\mu_t^{(1)}(X)}-\theta)|\F_{t-1}}
    &=\underset{\text{(i)}}{\underbrace{\mathbb{E}_{\hat P_t}[\widehat{\dot{\Psi}}(Z_t; \hat P_t)^2 \mid \mathcal{F}_{t-1}]}} 
+ \underset{\text{(ii)}}{\underbrace{\mathbb{E}_{\hat P_t}\left[\left(\mathbb{E}_{X \sim q(x)}[\hat\mu_t^{(1)}(X)] - \theta\right)^2 \mid \mathcal{F}_{t-1}\right]}} 
+ \underset{\text{(iii)}}{\underbrace{2 \mathbb{E}_{\hat P_t}\left[\widehat{\dot{\Psi}}(Z_t; \hat P_t)\left(\mathbb{E}_{X \sim q(x)}[\hat\mu_t^{(1)}(X)] - \theta\right) \mid \mathcal{F}_{t-1}\right]}}.
\end{align*}
For the term (i) 
\begin{align*}
    &\mathbb{E}_{\hat P_t}[\widehat{\dot{\Psi}}(Z_t; \hat P_t)^2 \mid \mathcal{F}_{t-1}]
    \\&= \ex{\hat P_t}{\brk{\frac{\Delta_t q(X_t)}{\hat\pi_t(X_t)\hat p_t(X_t)}\prn{Y_t-\hat\mu_t^{(1)}(X_t)}+\frac{(\Delta_t-\hat\pi_t(X_t))q(X_t)}{\hat \pi_t(X_t)\hat p_t(X_t)}(\hat\mu_t^{(1)}(X_t) - \hat\tau_t(X_t,F_t))}^2\big\vert \F_{t-1}},
\end{align*}
note that $\ex{\hat P_t}{\frac{(\Delta_t-\hat\pi_t(X_t))q(X_t)}{\hat \pi_t(X_t)\hat p_t(X_t)}(\hat\mu_t^{(1)}(X_t) - \hat\tau_t(X_t,F_t))\big\vert \F_{t-1}}=0$. It follows that 
\begin{align*}
    % \label{eq:square_decomp1}
    &\mathbb{E}_{\hat P_t}[\widehat{\dot{\Psi}}(Z_t; \hat P_t)^2 \mid \mathcal{F}_{t-1}] \nonumber\\
    &=\ex{\hat P_t}{\frac{(Y_t-\hat\mu_t^{(1)}(X_t))^2q(X_t)^2}{\hat\pi_t(X_t)\hat p_t(X_t)^2}+\frac{(1-\hat\pi_t(X_t))q(X_t)^2}{\hat p_t(X_t)^2}(\hat\mu(X_t)-\hat\tau_t(X_t,F_t))^2\big\vert \F_{t-1}}.
\end{align*}
We now introduce two lemmas that allows us to give asymptotic convergence of the above sequence.
\begin{lemma}[\cite{loeve1977elementary} p.165]\label{lem:convergence}
    Suppose $\{W_t\}_{t=1}^\infty$ are random variables with probability measure $P$ and $w$ be a constant. Let $0 < r < \infty$, suppose that $\mathbb{E}[|W_t|^r] < \infty$ for all $t$ and that $W_t \overset{P}{\to} z$ as $n \to \infty$. The following are equivalent:
\begin{itemize}
    \item[(1)] $W_t \to w$ in $L^r$ as $t \to \infty$;
    \item[(2)] $\mathbb{E}[|W_t|^r] \to \mathbb{E}_P[|w|^r] < \infty$ as $t \to \infty$;
    \item[(3)] The sequence $\{|W_t|^r,t\ge1\}$ is uniformly integrable, i.e., $\forall \epsilon >0,$ there exists $c>0$ such that $\ex{}{|W_t|^r\mathds 1\{|W_t|^r\ge c\}}<\epsilon$.
\end{itemize}
\end{lemma}
\begin{lemma}[\cite{hamilton2020time} Prop 7.7]
    Let $W_t, Z_t \in \mathbb{R}$ be random variables. Let $P$ be a probability measure of $Z_t$.
\begin{itemize}
    \item[(1)] Suppose there exist $r > 1$ and $M < \infty$ such that $\mathbb{E}_P[|W_t|^r] < M$ for all $t$. Then $\{W_t\}$ is uniformly integrable.
    \item[( 2)] Suppose there exist $r > 1$ and $M < \infty$ such that $\mathbb{E}_P[|Z_t|^r] < M$ for all $t$. If $W_t = \sum_{j=-\infty}^{\infty} h_j Z_{t-j}$ with $\sum_{j=-\infty}^{\infty} |h_j| < \infty$, then $\{W_t\}$ is uniformly integrable.
\end{itemize}
\end{lemma}
By the assumption, we have
\begin{align*}
    &\frac{(Y_t-\hat\mu_t^{(1)}(X))^2q(X)^2}{\hat\pi_t(X)\hat p_t(X)^2}+\frac{(1-\hat\pi_t(X))q(X)^2}{\hat p_t(X)^2}(\mu(X)-\tau_t(X,F_t))^2
    \\&\overset{a.s.}{\to} \frac{(Y-\mu(X))^2q(x)^2}{\pi^*(X)p^*(X)^2}+\frac{(1-\pi^*(X))q(X)^2}{ p^*(X)^2}(\mu(X)-\tau(X,F))^2
\end{align*}
Note that the left term is squared sum of sub-Gaussian random variables condition on $\F_{t-1}$ and hence is conditional sub-exponential. It follows that the left term is uniformly integrable and
\begin{align}
\label{eq:quad_decomp1}
    &\mathbb{E}_{\hat P_t}[\widehat{\dot{\Psi}}(Z_t; \hat P_t)^2 \mid \mathcal{F}_{t-1}] \to \tau^*.
\end{align}
For the term (ii), we similarly have 
\begin{align}
\label{eq:quad_decomp2}
    \ex{}{(\ex{X\sim q(x)}{\hat\mu_t^{(1)}(X)}-\theta)^2|\F_{t-1}} \to 0.
\end{align}
For the term (iii), we can get the convergence similarly
\begin{align}
\label{eq:quad_decomp3}
    &\ex{}{\mathbb{E}_{\hat P_t}\left[\widehat{\dot{\Psi}}(Z_t; \hat P_t)\left(\mathbb{E}_{X \sim q(x)}[\hat\mu_t^{(1)}(X)] - \theta\right) \mid \mathcal{F}_{t-1}\right]}\nonumber\\
    & = \ex{}{\ex{\hat P_t}{\frac{q(X_t)}{\hat p_t(X_t)}(\mu(X_t)-\hat\mu_t^{(1)}(X_t))(\ex{X\sim q(x)}{\hat\mu_t^{(1)}(X)}-\theta)|\F_{t-1} }}\nonumber\\ 
    & = \ex{}{(\ex{X\sim q(x)}{\hat\mu_t^{(1)}(X)}-\theta)^2}\to 0.
\end{align}
Combining \eqref{eq:quad_decomp1} \eqref{eq:quad_decomp2} and \eqref{eq:quad_decomp3} yields (a) in Lemma \ref{lem:clt}.

\paragraph*{\textbf{Verification of condition (b)}.} Since $xi_t$'s are sub-Gaussian, it follows directly from \cite{vershynin2018high}[Prop 2.5.2 (ii)].

\paragraph*{\textbf{Verification of condition (c)}.}
Define $d_t$ as $d_t = \xi_t^2-\ex{\hat P_t}{\xi_t^2|\F_{t-1}}$. Then $\{d_t\}_{t=1}^\infty$ forms a martingale. Since $d_t$ is bounded by the assumption, the weak law of large numbers for martingale \cite{hall2014martingale}[Sec 2.5] guarantees that 
\begin{align*}
    \frac{1}{T}\sum_{t=1}^Td_t = \frac{1}{T}\sum_{t=1}^T(\xi_t^2-\ex{\hat P_t}{\xi_t^2|\F_{t-1}} \overset{p}{\to} 0.
\end{align*}
Following a similar streamline in verification of condition (a) and the proof of Lemma 10 in \cite{hadad2021confidence}, we have 
\begin{align*}
    \frac{1}{T}\sum_{t=1}^T\ex{\hat P_t}{\xi_t^2|\F_{t-1}}\overset{p}{\to} 1.
\end{align*}
It follows that 
\begin{align*}
    \frac{1}{T}\sum_{t=1}^T\xi_t^2 = \frac{1}{T}\brk{(\xi_t^2-\ex{\hat P_t}{\xi_t^2|\F_{t-1}})+\ex{\hat P_t}{\xi_t^2|\F_{t-1}} }\overset{p}{\to}1.
\end{align*}

\subsection{Proof of Lemma \ref{lem:martingale}}\label{sec:proof_lemma}
First note that 
\begin{align*}
   \widehat{\dot{\Psi}}(Z_t;\hat P_t):=   \frac{\Delta_t q(X_t)}{\hat\pi_t(X_t)\hat p_t(X_t)}\prn{Y_t-\hat\mu_t^{(1)}(X_t)}+\frac{(\Delta_t-\hat\pi_t(X_t))q(X_t)}{\hat \pi_t(X_t)\hat p_t(X_t)}(\hat\mu_t^{(1)}(X_t) - \hat\tau_t(X_t,F_t)).
\end{align*}
Condition on $X_t$, we know that $\ex{}{\Delta_t|X_t}=\hat\pi_t(X_t)$. Thus the second term above equals to zero under expectation. For the first term, we take expectation sequentially and get 
\begin{align*}
    \ex{\hat P_t}{\hatpsi(Z_t;\hat P_t\vert \F_{t-1}} = \ex{X\sim q(x}{\hat\mu_t^{(1)}(X)-\mu(X)\vert \F_{t-1}}.
\end{align*}
Substituting it into $\ex{Y_t\sim \P(y|X_t),X_t\sim \hat p_t(x),\Delta_t\sim \hat\pi_t(x)}{\xi_t|\F_{t-1}}$ leads to the result.

% \subsection{Proof of Theorem \ref{thm:efficient2}}\label{sec:proof_efficient2}
% We follow the proof in Appendix \ref{sec:proof_efficient}. With a similar streamline of deriving \eqref{eq:decomp1}, we can get 
\end{APPENDICES}
%
%   or
%
% \begin{APPENDICES}
% \section{<Title of Section A>}
% \section{<Title of Section B>}
% etc
% \end{APPENDICES}

%%

%%%%%%%%%%%%%%%%%
\end{document}